\documentclass[10pt,twocolumn,letterpaper]{article}

\usepackage{iccv}
\usepackage{times}
\usepackage{epsfig}
\usepackage{graphicx}
\usepackage{amsmath}
\usepackage{amssymb}

\usepackage{booktabs}
\usepackage{subcaption}
\usepackage{multirow}
\usepackage[dvipsnames]{xcolor}
\definecolor{Pantone300C}{HTML}{0065BD}
\definecolor{MediumGray}{HTML}{808080}
\definecolor{Pantone158}{HTML}{E37222}
\usepackage[nolist,nohyperlinks]{acronym}
\makeatletter
\@namedef{ver@everyshi.sty}{}
\makeatother
\usepackage{tikz,pgfplots}
\usepackage{array,makecell}

\usepackage[breaklinks=true,bookmarks=false]{hyperref}

\iccvfinalcopy

\ificcvfinal\fi

\begin{document}
\begin{acronym}
\acro{AE}[AE]{Absolute Error}
\acro{APE}[APE]{Absolute Percentage Error}
\acro{BMI}[BMI]{Body Mass Index}
\acro{CCE}[CCE]{Categorical Cross-Entropy}
\acro{MSE}[MSE]{Mean Squared Error}
\acro{MAPE}[MAPE]{Mean Absolute Percentage Error}
\acro{MAE}[MAE]{Mean Absolute Error}
\acro{NHANES}[NHANES]{National Health and Nutrition Examination Survey}
\acro{SMPL}[SMPL]{Skinned Multi-Person Linear Model}
\end{acronym}

\title{VolNet: Estimating Human Body Part Volumes from a Single RGB Image}

\author{Fabian Leinen\\
Technical University of Munich\\
Audi AG\\
{\tt\small fabian.leinen@tum.de}
\and
Vittorio Cozzolino\\
Technical University of Munich\\
{\tt\small vittorio.cozzolino@in.tum.de}
\and
Torsten Schön\\
Technische Hochschule\\Ingolstadt\\
{\tt\small torsten.schoen@thi.de}
}

\maketitle
\ificcvfinal\thispagestyle{empty}\fi

\noindent
\begin{abstract}
Human body volume estimation from a single RGB image is a challenging problem despite minimal attention from the research community. However  VolNet, an architecture leveraging 2D and 3D pose estimation, body part segmentation and volume regression extracted from a single 2D RGB image combined with the subject's body height can be used to estimate the total body volume. VolNet is designed to predict the 2D and 3D pose as well as the body part segmentation in intermediate tasks. We generated a synthetic, large-scale dataset of photo-realistic images of human bodies with a wide range of body shapes and realistic poses called SURREALvols\footnote{\url{https://github.com/fleinen/SURREALvols}}. By using Volnet and combining multiple stacked hourglass networks together with ResNeXt, our model correctly predicted the volume in $\sim$82\% of cases with a 10\% tolerance threshold. This is a considerable improvement compared to state-of-the-art solutions such as BodyNet with only a $\sim$38\% success rate.

\end{abstract}

\begin{figure}[hbt!]
   \centering
   \resizebox{0.992\columnwidth}{!}{%
   \setlength\tabcolsep{1.5pt}
   \begin{tabular}{
   >{\arraybackslash}m{2.5cm}
   >{\centering\arraybackslash}m{1.8cm}
   >{\centering\arraybackslash}m{1.8cm}
   >{\centering\arraybackslash}m{1.8cm}
   }
        RGB Image &
        \includegraphics[width=1.8cm]{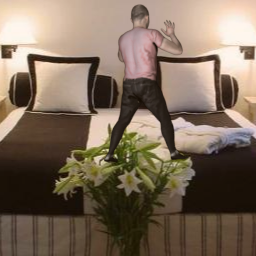}&
        \includegraphics[width=1.8cm]{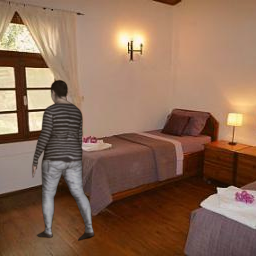}&
        \includegraphics[width=1.8cm]{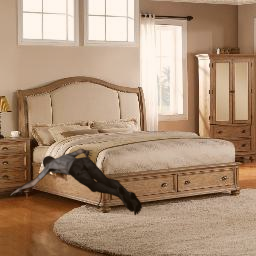}\\
        Body Height & 157 cm & 168 cm & 182 cm \\
        2D Pose &
        \includegraphics[width=1.8cm]{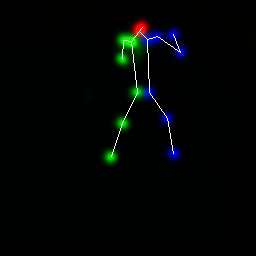}&
        \includegraphics[width=1.8cm]{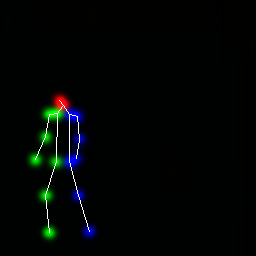}&
        \includegraphics[width=1.8cm]{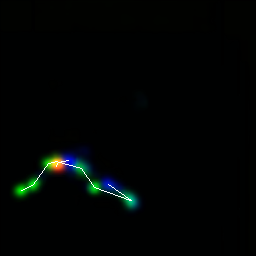}\\
        Segmentation &
        \includegraphics[width=1.8cm]{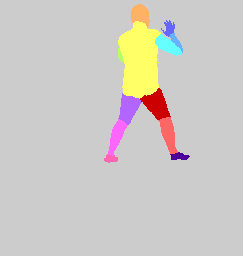}&
        \includegraphics[width=1.8cm]{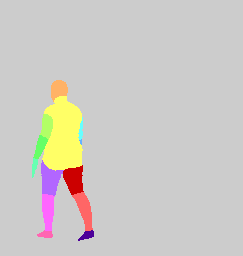}&
        \includegraphics[width=1.8cm]{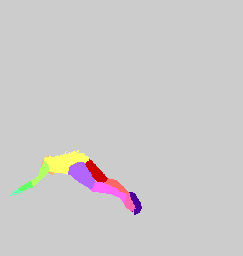}\\
        3D Pose &
        \includegraphics[width=1.8cm]{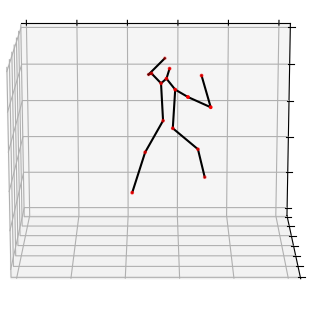}&
        \includegraphics[width=1.8cm]{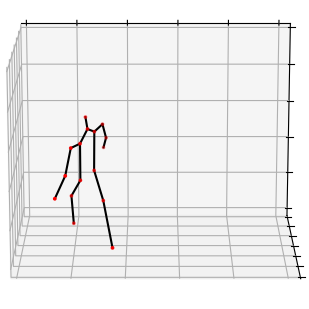}&
        \includegraphics[width=1.8cm]{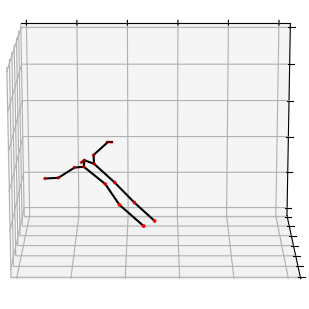}\\
        Head               & 7.97  {\scriptsize / 7.84}            & 7.75  {\scriptsize / 7.53} & 7.19  {\scriptsize / 6.65} \\
        Torso              & 42.74 {\scriptsize / 42.05}           & 54.24 {\scriptsize / 52.56}& 33.22 {\scriptsize / 26.72} \\
        Left Upper Arm     & 1.70  {\scriptsize / 1.70}            & 2.33  {\scriptsize / 2.28} & 1.50  {\scriptsize / 1.16} \\
        Left Fore Arm      & 1.50  {\scriptsize / 1.53}            & 1.60  {\scriptsize / 1.66} & 1.28  {\scriptsize / 1.18} \\
        Left Hand          & 0.64  {\scriptsize / 0.71}            & 0.66  {\scriptsize / 0.74} & 0.71  {\scriptsize / 0.83} \\
        Right Upper Arm    & 2.05  {\scriptsize / 2.02}            & 2.85  {\scriptsize / 2.63} & 1.65  {\scriptsize / 1.31} \\
        Right Fore Arm     & 1.45  {\scriptsize / 1.47}            & 1.63  {\scriptsize / 1.70} & 1.40  {\scriptsize / 1.26} \\
        Right Hand         & 0.65  {\scriptsize / 0.75}            & 0.66  {\scriptsize / 0.73} & 0.70  {\scriptsize / 0.81} \\
        Left Up Leg        & 5.79  {\scriptsize / 5.18}            & 7.30  {\scriptsize / 7.11} & 5.72  {\scriptsize / 3.89} \\
        Left Lower Leg     & 2.83  {\scriptsize / 2.56}            & 3.57  {\scriptsize / 3.43} & 3.02  {\scriptsize / 2.07} \\
        Left Foot          & 1.43  {\scriptsize / 1.43}            & 1.75  {\scriptsize / 1.88} & 1.82  {\scriptsize / 1.72} \\
        Right Up Leg       & 5.73  {\scriptsize / 5.21}            & 7.29  {\scriptsize / 7.05} & 5.47  {\scriptsize / 3.55} \\
        Right Lower Leg    & 2.79  {\scriptsize / 2.54}            & 3.57  {\scriptsize / 3.45} & 2.97  {\scriptsize / 2.11} \\
        Right Foot         & 1.47  {\scriptsize / 1.44}            & 1.75  {\scriptsize / 1.83} & 1.89  {\scriptsize / 1.78} \\
        \textbf{Total}     & \textbf{78.72  {\scriptsize / 76.41}} & \textbf{96.95 {\scriptsize  / 94.57}} & \textbf{68.55  {\scriptsize / 55.04}}
    \end{tabular}}
    \caption{Two good and one bad predictions with all inputs, outputs and intermediate results. The volumes are given in $dm^3$. In each case, the first value represents the predicted value, the second one the ground truth value.}
    \label{fig:results_examples_case}
\end{figure}

\section{Introduction}
\noindent
Reconstruction of 3D poses and shapes from a single 2D image has received increasing attention from the deep learning community \cite{Bogo_ECCV_2016,Zanfir_2018_CVPR,Pavlakos_2018_CVPR,Kolotouros_2019_ICCV,Luo_2020_CVPR,Wu_2020_CVPR,Chen_2020_CVPR}. Currently, limited work has been done on estimating the volume of objects and, especially, human body parts. In fact, similar work is limited to estimating the shape of an object e.g. by estimating the parameters of a model similar to \ac{SMPL}~\cite{SMPL2015}. However, there are many applications in the fields such as ergonomics, virtual try on, and medicine, where body volume fractions are crucial. Furthermore, the weight of individual body parts can be calculated using volume and density. Our study shows that the volume of body parts can be estimated from a single 2D RGB image. To do this, we designed and implemented a novel custom neural network architecture based on the stacked hourglass \cite{newell2016stacked} approach. Our main contributions are the following:
\begin{itemize}
\item We extended the SURREAL dataset~\cite{varol17_surreal} with volumes of 14 different body parts leading to a novel benchmark dataset called \textbf{SURREALvols}.
\item We introduced a novel network architecture that can estimate the total volume of a human being as well as the volumes of 14 individual body parts using a single RGB image with an accuracy of \textbf{$\sim$82\%} with a tolerance threshold of 10\,\% of the body volume.
\item We show how to improve the volume regression for single body parts when fine-tuning the network, which leads to an additional $\sim$2\% increase in \ac{MAPE}.
\end{itemize}

\section{Related Work}
\label{sec:related_work}

\textbf{Pose and 3D Body Representation.}
Estimating human pose from an individual 2D image is possible by leveraging neural networks based on a stacked hourglass architecture~\cite{Toshev_2014_CVPR,newell2016stacked,Gueler_2018_CVPR,Andriluka_2014_CVPR}. Typically, multiple hourglass modules are stacked together enabling the network to reevaluate initial guesses. Next to the pose, there are many applications requiring a digital 3D representation of the human body. For example, VR and AR~\cite{Augmented_Reality2010,Habermann_2020_CVPR}, computer animations~\cite{Badler1997,Thalmann1996,Mori2006}, or in car monitoring.
\ac{SMPL} is a generative human body model trained over real body scans~\cite{SMPL2015}. The shape $\beta \in \mathbb{R}^{10}$ is obtained from the first coefficients of the PCA while the pose $\theta \in \mathbb{R}^{3K}+3$ is modeled by the relative rotations of the $K=23$ joints in axis-angle representation. Overall, \ac{SMPL} provides a male and female template mesh, each consisting of 6890 vertices and 23 joints. Thus, a considerable number of realistic human body shapes can be simulated in different poses.

The dataset is organized into a maximum of 100 frame clips where, body shape, texture, background, lighting, and camera position are constant, and only the person's pose and location change. Subjects are based on \ac{SMPL} which allow generation of people with a wide range of poses and shapes. To ensure that the poses and motions look realistic, positions are taken from the CMU MoCap database~\cite{mocap} which contains 3D positions of MoCap markers of more than 2000 sequences. Fitting \ac{SMPL} to these markers leads directly to the \ac{SMPL} pose parameters. At the beginning of each clip, the person is placed at the center of the picture and is allowed to move away within one clip. \ac{SMPL}'s ten shape parameters are sampled randomly and not changed within one clip. To achieve a photo-realistic appearance of the resulting RGB data, a human texture is added to the mesh. Textures were extracted from CAESAR~\cite{Robinette_caesar_2002} scans showing people in tight-fitting as well as usual clothing. The textured person is rendered in front of a real-world photo randomly sampled from the LSUN dataset~\cite{Yu_LSUN_2015} using only images from the kitchen, living room, bedroom, and dining room categories. The background and the person have no relation to each other, leading to people floating and intersecting with invisible objects. To further improve variance, camera pose and lighting were randomly sampled.\par

\textbf{3D reconstruction.}
Reconstructing the 3D position and 3D shape of arbitrary and specific objects has received increasing attention. Navaneet \etal directly regressed 3D features using a deep neural network from a 2D image~\cite{navaneet2019differ}. For the purpose of estimating the 3D shape  of a human body, the typical technique is to regress the parameters of a predefined human shape model such as \ac{SMPL}.
Bogo \etal \cite{BogoSMPLify2016} extracted keypoints from RGB images and fit them to \ac{SMPL}. Lassner \etal extended this approach to use 2D silhouettes \cite{LassnerUP2017}. BodyNet~\cite{varol18_bodynet} represents the 3D body in the form of voxels, different from previous approaches directly estimating the 3D body pose and shape. The BodyNet architecture uses multiple stacked hourglass networks to predict the 2D and 3D pose, as well as the body part segmentation. Before estimating the actual 3D shape in a voxel grid. The output is then optionally fit to the \ac{SMPL} model. BodyNet is trained on the SURREAL~\cite{varol17_surreal} and Unite the People~\cite{LassnerUP2017} datasets.
Xu \etal used DensePose~\cite{guler2018densepose} to estimate an IUV map as a proxy representation to fit \ac{SMPL}~\cite{Xu_2019_ICCV}. To improve the training process, they use a \textit{render-and-compare} approach.
Alldieck \etal proposed an approach that predicts 3D human shapes in a canonical pose from an input video showing the subject from all sides \cite{alldieck2018detailed}. A deep convolutional network predicted the shape parameters only. Like Xu \etal, they improved their training process using \textit{render-and-compare} on a few frames. In subsequent work they introduced Tex2Shape to generate a detailed 3D mesh from only a single RGB image~\cite{alldieck2019tex2shape}, and made use of DensePose's  UV coordinates to first regress \ac{SMPL} shape parameters $\beta$ using a simple convolutional network architecture and then generated normals from the DensePose's texture maps using GAN architecture. The normals were then used to refine the 3D mesh, resulting in a detailed 3D mesh of people wearing clothes.

\textbf{Visual Body Weight Estimation.}
Velardo \etal estimated human body weight using a linear regression model that maps seven different body measurements (body height, upper leg length, calf circumference, upper arm length, upper arm circumference, waist circumference, and upper leg circumference) to the body weight~\cite{velardo2010weight}. Data from \ac{NHANES}~\cite{nhanes} was used for optimization. Due to the lack of data for human bodies with total body weight below 35 Kg or above 130 Kg, consequently only 27k  data points were available. The resulting model predicted the weight of 60\,\% of samples from the test set with a maximum error of $\pm5\,\%$. 93\,\% of the samples were predicted with a maximum error of $\pm10\,\%$. Since this model is biased because it was constructed with data collected only in the USA.
\par
To estimate the weight from 3D objects, Jia \etal \cite{Jia2012} established a mathematical model to calculate the volume of food from 2D images. Then Xu \etal \cite{Xu2013} improved the food volume estimation based on prior knowledge. In 2016, Hippocrate \etal \cite{Hippocrate2016} demonstrated how to not only derive the volume of food, but directly estimate its weight. While for human beings, most publications concentrate on human shape estimation \cite{Tong2012,Gabeur_2019_ICCV,Zhang_2017_CVPR}, with little work directly estimating weight or body mass index \cite{dantcheva2018,Affuso2018}.

\section{Baseline Algorithm}
\begin{figure}[tb]
\centering
    \begin{tikzpicture}
    \begin{axis}[
        legend pos=south east,
		grid=major,
        xlabel={\ac{MAPE}},
        ylabel={Ratio within \ac{MAPE}},
        ylabel near ticks,
        width=\linewidth,
        height=4cm,
        xmin = 0,
        xmax = 100,
        ymin = 0,
        ymax = 1,
    ]
	
	\addplot[ultra thick, mark=none] table [x=rel, y=cum, col sep=comma]{data/baseline/cum_bodynet_testset.csv};
	
    \end{axis}
    \end{tikzpicture}
\caption{Cumulative distributions of the \ac{MAPE} of BodyNet. This graph shows the ratio of correctly predicted samples (ordinate) using a certain threshold (abscissa).}
\label{fig:BaselineAlgorithms}
\end{figure}
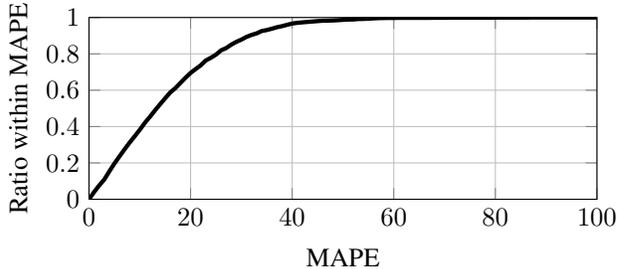

\noindent
Even thought Alldieck~\etal achieved visually good results with Tex2Shape~\cite{alldieck2019tex2shape}. Their approach aimed to predict a 3D mesh including clothes, hair and wrinkles, in the present study we aimed to predict body volumes in the absence of clothing. Moreover, The data used by Allideck \etal is not publicly available making it impossible to calculate ground truth volumes for comparison \cite{alldieck2019tex2shape}.
Therefore, while BodyNet's output of a voxel grid without the person's body height. and requires manual volume calculations, we used this as the baseline algorithm for our approach. However, for our evaluation, the test set of SURREALvols was used as the input for BodyNet. BodyNet is trained on SURREAL, which uses the same poses and textures with the same training and test split, making it a reasonable comparison for the following approaches. Also, BodyNet expects input images cropped to the bounding box of the person, so the images from SURREALvols were cropped and resized. Then, for each input image, BodyNet returned a $128 \times 128 \times 128$ probability grid. This grid was converted to a binary voxel grid using a threshold of $0.5$ as reported in the official BodyNet implementation\footnote{\url{https://github.com/gulvarol/bodynet}}.
As previously mentioned, BodyNet does not contain real volume information since it is not scaled with respect to absolute references. Thus, first the edge length of a single voxel is calculated by using the highest and the lowest points of the predicted person. By subtracting the lowest point from the highest, an absolute height is obtained. It is important to note that, since points are chosen in the current pose, the highest point may be a lifted hand which could lead to possible errors in the estimated value for body height. This was done for the ground truth mesh, resulting in height in meters $h_m$, and for the predicted voxel grid, resulting in height in numbers of voxels $h_v$. Thus, the reference points can be easily obtained for both the ground truth model and BodyNet's prediction. Using these two values, the edge length $l_q$ and the volume of a single voxel $V_q$ were obtained. To calculate the total volume of BodyNet's prediction $V_{tot}$, the total number of filled voxels $|\text{Voxels}|$ is used:
\begin{align}
    l_q &= \frac{h_m}{h_v}\\
    V_q &= l_q^3\\
    V_{tot} &= V_q \cdot |\text{Voxels}|
\end{align}
The total volume of BodyNet's output $V_{tot}$ was compared to the ground truth volume, where the mean \ac{AE} was 11.8\,\% with $\sigma=9.82$ and the \ac{APE} mean is 15.66\,\% with $\sigma=11.76$ measuring in $dm\textsuperscript{3}$. The cumulative distribution of the \ac{APE} is visualized in Figure \ref{fig:BaselineAlgorithms}.

\label{section:BaselineAlgorithms}

\section{Approach}
\label{sec:approach}
\noindent
Our approach can be divided in two parts: data generation and model training.
Our newly generated dataset was based on SURREAL~\cite{varol17_surreal}. We calculates the volume of 14 body parts and used the body height obtained in the model's neutral pose as a reference.
This data was then used as ground truth information.
The learning goal of the network architecture was to estimate the individual body part volumes by using a single RGB image as well as the body height. Due to information loss when projecting a 3D representation onto a 2D image plane, estimating the volume without knowing the body height is non-trivial.
Even though there are multiple approaches to estimate the body height by correlating different reference lengths \cite{BenAbdelkaderStatBodyHeight}, it is more promising to estimate the body height from well-known keypoints extracted from the person's surroundings.
Unless stated differently, in our approach body height is considered a known parameter and is used as an additional input for our model. In the ablation studies, we report results with unknown body height.
Our proposed model predicts the 2D and 3D pose as well as the body part segmentation before finally regressing the subjects body part volumes.

\subsection{Data}
\label{ssec:data}
\noindent
Ideally, a dataset that can be used for body volume estimation consists of photo-realistic images of humans in realistic poses combined with a wide range of different labels, such as 2D and 3D poses, body part segmentation, body height, and body volume. Most of these attributes are available in the SURREAL dataset except for body height. To create SURREALvols, we extended the SURREAL dataset. While the basic structure of SURREAL remained unchanged, we generated RGB images and body part segmentations. Moreover, the same poses, textures, and training/validation split were used as in ~\cite{varol17_surreal}. Backgrounds were randomly sampled from bedroom images within the LSUN dataset~\cite{Yu_LSUN_2015}. Since this category alone contains roughly 3 million images, the variance in backgrounds is sufficient for good results. Due to a reported error in SURREAL's original mapping of mesh vertices to segments, the segmentation may contain some errors\footnote{https://github.com/gulvarol/surreal/issues/7}.
Even though the impact of this bug is quite low, it has been fixed for SURREALvols. First, the RGB image and segmentation masks are generated. Next, we calculated the 2D and 3D pose annotations, and finally the shape parameters were used to determine the person's body height and body part volumes. For calculating the body height, the shape parameters were applied in \ac{SMPL}'s neutral pose. Then, we determined the highest $\vec{p_h}$ and lowest $\vec{p_l}$ points and body height was calculated by subtracting their y-values, where y is the axis pointing to the top.
Instead of calculating the overall body volume, we calculated the volume of each body part. This more detailed representation is required for some use cases and can help in improving the learning process of the model by breaking it down into a subset of simpler tasks. To calculate the volume of each body part, \ac{SMPL}'s mesh was split into the corresponding body parts.
The mapping of vertices to body parts was given by SURREAL when generating the 2D segmentation where the incorrectly mapped vertices were corrected. It is noteworthy that in the given mapping, the fingers and toes do not belong to the hands and feet. Also the torso is split into seven parts. For consistency with BodyNet's segmentation network, these parts are merged to represent the same fourteen logically reasonable body parts. In order to calculate a mesh's volume, we manually preprocessed it to ensure that they were manifold.
The result is a list for each body part containing the mesh' faces that consists of vertices from the \ac{SMPL} model, referred to as template. To calculate the volumes of the body parts, \ac{SMPL}'s pose and shape parameters were applied, and changed the position of the vertices. Next, the volume was calculated as:
\begin{align}
	\begin{split}
		V_{tot} &= \sum_{i=0}^{|\text{faces}|} \frac{1}{6}\cdot \vec{f_{i1}} \cdot  \vec{f_{i2}} \times \vec{f_{i3}}\\
		\label{eq:MeshVolume}
	\end{split}
\end{align}
where $|\text{faces}|$ denotes the total number of faces in the mesh and $f_{i1}$ to $f_{i3}$ are the three vertices of a single face $f_i$.\par
\begin{figure}[!tb]
	\centering
	\begin{tabular}{c@{\hskip 4pt}c@{\hskip 4pt}c@{\hskip 4pt}c@{\hskip 4pt}c@{\hskip 4pt}}
		\includegraphics[width=1.5cm]{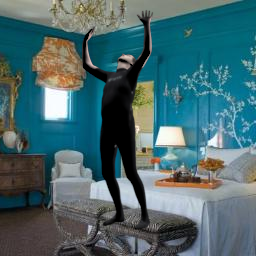}&
		\includegraphics[width=1.5cm]{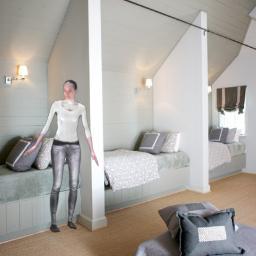}&
		\includegraphics[width=1.5cm]{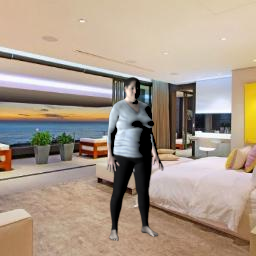}&
		\includegraphics[width=1.5cm]{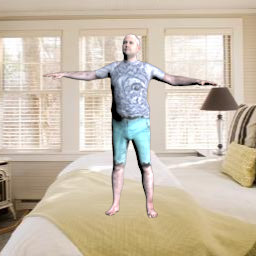}&
		\includegraphics[width=1.5cm]{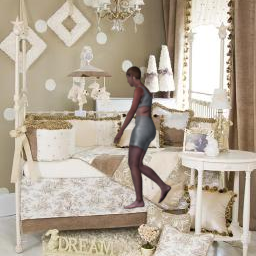}\\
		\includegraphics[width=1.5cm]{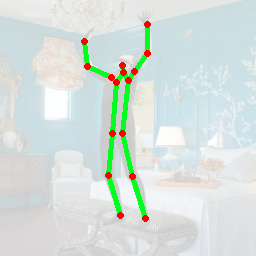}&
		\includegraphics[width=1.5cm]{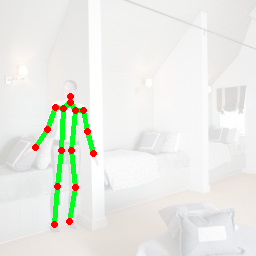}&
		\includegraphics[width=1.5cm]{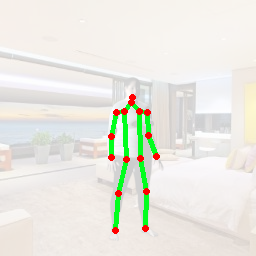}&
		\includegraphics[width=1.5cm]{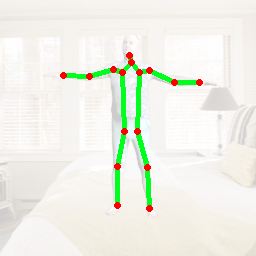}&
		\includegraphics[width=1.5cm]{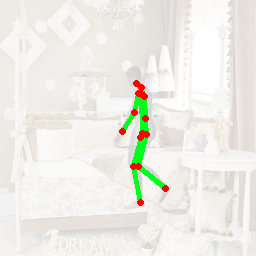}\\
		\includegraphics[width=1.5cm]{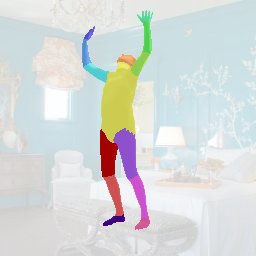}&
		\includegraphics[width=1.5cm]{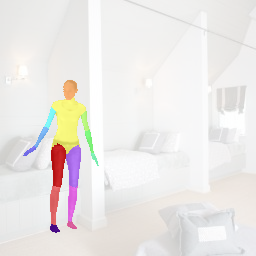}&
		\includegraphics[width=1.5cm]{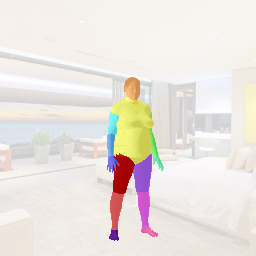}&
		\includegraphics[width=1.5cm]{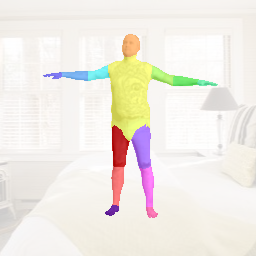}&
		\includegraphics[width=1.5cm]{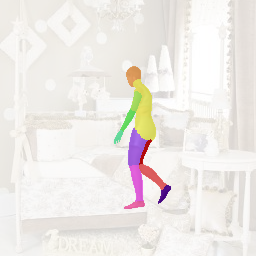}
	\end{tabular}
	\caption{Example images from SURREALvols showing only subjects that are fully visible. The raw image (first row), the 2D pose (second row) and the body part segmentation (last row) are visualized.}
	\label{fig:gensurreal8examples}
\end{figure}
Following this procedure, a dataset contained 372,142 RGB images each with a resolution of 256 px $\times$ 256 px. Additionally, the 2D and 3D poses, as well as the body part segmentations were generated. The 2D and 3D poses consisted of 23 joints and the body part segmentation consisted of 25 segments including the background. Examples are shown in Figure \ref{fig:gensurreal8examples}, where some joints are omitted and some segments are merged together for illustration purposes. The structure corresponds to SURREAL, where each image, or frame, is assigned to a group, or clip, containing up to 100 images. For SURREALvols, the background image, the lighting, the distance to the camera, the rotation about the person's vertical axis, the person's gender and shape were all fixed. Fixing the shape for all images within a clip resulted in the same body part volumes within one clip. The location of the person, represented as $u$ and $v$ coordinates in image space, was sampled for each frame. Additionally, the pose was changed using the next pose from a CMU MoCap sequence, as done in~\cite{varol18_bodynet}. The clips were split in to training, validation and test sets, so that all the frames in one clip belonged to either one of the sets. For the training and validation sets, poses, textures, and background images were randomly sampled. The test set was generated using new backgrounds and randomly sampled poses and textures that were used in the training and validation sets. Figure \ref{tab:gen_surreal8Results} shows further statistics of the generated data.
\begin{table}[h]
	\centering
	\begin{tabular}{@{}lrrr@{}}
		\toprule
		& train   & validation & test       \\ \midrule
		clips                & 9,298   & 2,255      & 2,500       \\
		frames total         & 297,482 & 72,160     & 2,500       \\
		frames fully visible & 194,474 & 48,698     & 2,014       \\ \bottomrule
	\end{tabular}
	\caption{Number of frames contained in SURREALvols}
	\label{tab:gen_surreal8Results}
\end{table}
\noindent
The last row lists the number of frames where the person is fully visible. For validation, only one frame with a fully visible person per clip was used. One clip had no frame where the person was fully visible, and only 2254 frames were used for validation.\par
The mean body volume in the training set was 79.4 dm\textsuperscript{3} where the torso accounted for the largest part with 42.1 dm\textsuperscript{3}, followed by the head with 7.4 dm\textsuperscript{3}, the upper legs with 5.9 dm\textsuperscript{3} and 5.8 dm\textsuperscript{3} on the right and left side, respectively. Furthermore, the body heights were sampled to follow two individual Gaussian distributions, one for each gender. The sampling process ensured a wide range of attributes while still being comparable for training, validation and test sets.\par

\subsection{Model Training}
\label{ssec:Network}

\begin{figure*}[tb]
	\centering
	\includegraphics[width=1\textwidth]{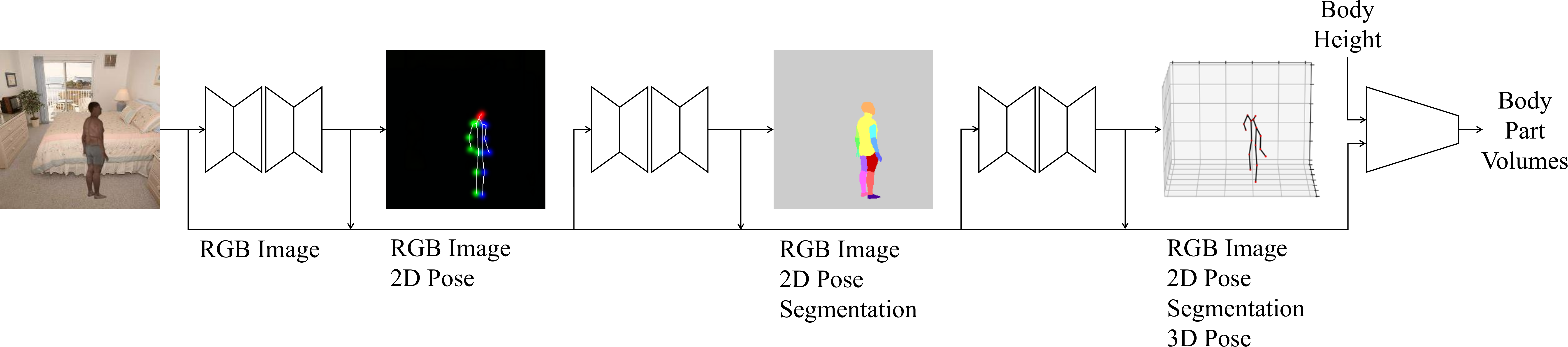}
	\caption{Architecture of VolNet. The 2D pose is predicted from a single RGB image. Next, the RGB image and the 2D pose are used as input for the body part segmentation. Together these predict the 3D pose. All four intermediate results together combined with the body height are then used to predict the volume of each individual body part.}
	\label{fig:VolNetArch}
\end{figure*}

\noindent
VolNet goes through multiple intermediate tasks before regressing the volume of the body parts. As shown in this section, these tasks help the network to gain a deeper understanding of the scene. First, the 2D pose (the pixel locations of the 16 joints) is estimated and acts similar to an attention map for the subsequent segmentation step. Further, it provides the first indication of the perspective.
In the next step, the image is segmented into the 14 body parts plus the background. The surface areas provide essential information about the size and volume of each body part and is highly dependent on the real 3D pose of the subject. For example, if a person stands sideways to the camera, the calculated surface is far smaller compared to a subject standing face-forward. 
The 3D pose is estimated during the last intermediate tasks. Combined with the raw RGB image, the output of each subtask is used in the final volume regression network to estimate the subject's body part volumes. The tasks build upon each other. While the 2D pose estimation only uses the raw RGB image as input, the body part segmentation uses the findings of both the 2D pose estimation and the raw RGB image. The 3D pose estimation then uses the RGB image as well as the estimated 2D pose and body part segmentation. The overall architecture is shown in Figure \ref{fig:VolNetArch}. Each intermediate tasks adds more knowledge and improves the learning process. Additionally, they can be considered domain-invariant feature representations which make domain adaption even easier~\cite{pmlr-v28-muandet13}.\par
In theory, an unlimited amount of data for SURREALvols can be generated. In practice, however, the number of certain variance factors, especially textures, is limited. To boost the ability to generalize, the RGB images are augmented by manipulating the color, contrast and brightness, adding random noise~\cite{moreno2018noiseadjustment}, and blurring the image during training~\cite{Shorten2019ASO}.

\subsubsection{Intermediate Tasks}

\begin{figure*}[tb]
	\centering
	\includegraphics[width=1\textwidth]{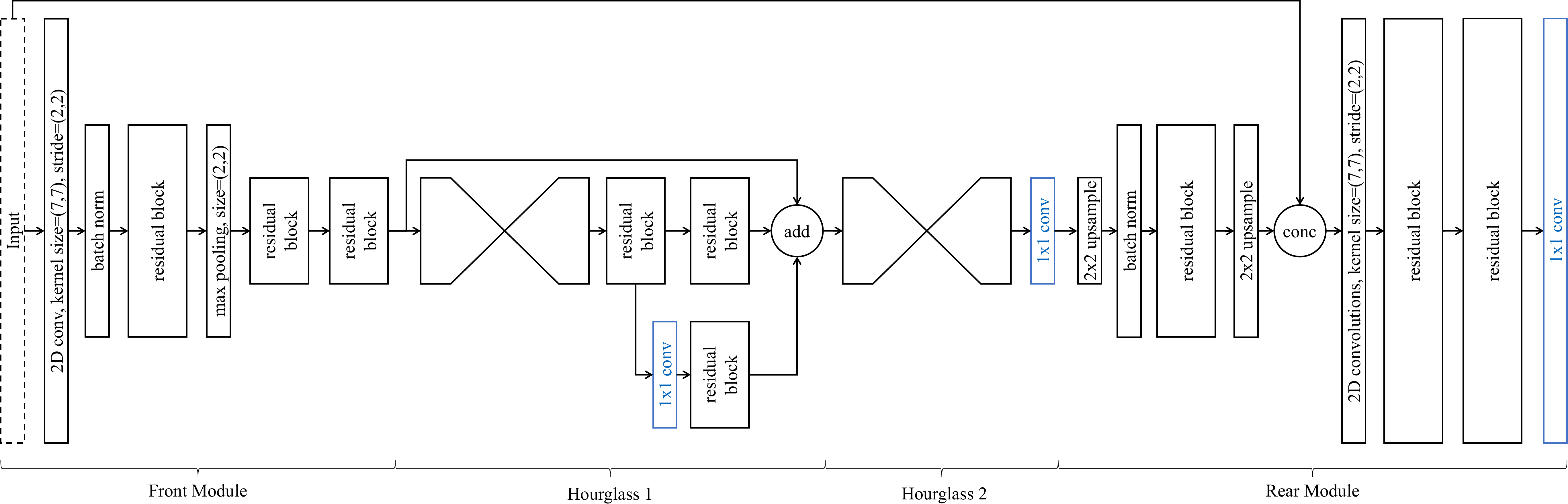}
	\caption{Structure of each of the first three subnetworks. The front module and both hourglasses are adapted from Newell \etal~\cite{newell2016stacked}. Final results from the last stack are upsampled to 256~$\times$~256 again. Losses are applied after the blue blocks.}
	\label{fig:s2hg}
\end{figure*}

All intermediate tasks use a stacked hourglass architecture \cite{newell2016stacked} which uses 256~$\times$~256 inputs and produces 64~$\times$~64 outputs. For consistency, we set the spatial resolution of the input and output to 256~$\times$~256. The entire architecture of a single subnetwork is shown in Figure \ref{fig:s2hg}. The front module is responsible for decreasing the spatial resolution to 64~$\times$~64, and both stacks are taken from the original stacked hourglass network.
In order to balance performance and inference time, every subnetwork uses only two stacks of hourglasses. We extended this architecture with a rear module that increased the spatial resolution to 256~$\times$~256 again. This module combined 2~$\times$~2 nearest-neighbor scalings, batch normalization~\cite{ioffe2015batch}, and residual blocks. After reaching a resolution of 256~$\times$~256, the input is concatenated to allow the network to correct the predictions on a pixel-level. The loss is calculated for the output of each stack in 64 $\times$ 64 and also for the final output of the layer in $256 \times 256$.\par
The subnetwork for estimating the 2D pose is structured as described above where only a single RGB image was used as input. The output was a tensor with 16 channels, one for each joint. SURREALvols contained 23 different joints but only those considered essential to describe the body pose were selected\footnote{head, neck, left + right shoulder, left + right arm, left + right fore arm, left + right hand, left + right hip joint, left + right knee, left + right foot}.
RMSprop optimizer with a learning rate of $2.5*10^{-4}$ and \ac{MSE} loss was used. The described subnetwork correctly predicted 97.00\% of all keypoints, using the PCK@0.05 metric~\cite{yang_2013_pck_metric}. With 89.04\%, the network performed worst for the right hand. It performed best for the head with 99.87\% of the samples classified correctly.\par

For inferring the body part segmentation, the raw RGB image was used as well as the predicted 2D pose. Both are concatenated in their channel dimension resulting in an input tensor with 19 channels. The output of this network had 15 channels, 14 different classes of body parts plus one for the background. For training, \ac{CCE} loss and Adam~\cite{Kingma_2014_adam} with default parameters and learning rate $\alpha = 2.5*10^{-4}$ are used. The overall performance measured in IoU (excl. BG) on the validation set was 81.17\%. The best values were achieved for the head and torso, with 92.51\% and 92.32\% respectively. In contrast, the performance on the hands was the worst with 69.02\% and 69.84\%. For the background, we reached an IoU of 99.73\%.\par

The subnetwork responsible for estimating the 3D pose used as inputs the RGB image, the 2D pose, and the body part segmentation which were concatenated along their channel dimension resulting in a tensor with 34 channels. As in BodyNet, the 3D pose is represented as a 16 voxel grid, one for each joint, containing a multivariate Gaussian with its mean at the position of the joint. BodyNet assumes that the depth of a body roughly fits in 85~cm, split into 19 bins where each bin represents a range of 5~cm. For our approach, we instead used the relative depth in an orthogonal projection and split the depth into 12 bins rather than 19.

This subnetwork mainly focused on estimating the depth value for each joint since the spatial position was already known from the first subnetwork. As with the 2D pose estimation, this subnetwork was trained using RMSprop optimizer with a learning rate of $2.5*10^{-4}$ and \ac{MSE}. For evaluation, the accuracy with a fixed threshold of 12 pixels for the spatial resolution and two bins for depth were used. The maximum accuracy of 82.55\% was reached.

\subsubsection{Volume Regression}

The last step in VolNet predicts the volume of 14 predefined body parts. It uses the RGB image combined with the results from the previously mentioned subtasks concatenated along their channel dimension. This results in a single $256\times256\times226$ tensor. The body height in centimeters is used as an additional input to the network. Hence, the subnetwork must deal with spatially dependent data as well as single scalar data. 
For the convolutional component, we used a standard ResNeXt-50~\cite{xie2016aggregated} with a cardinality of 32. The body height is concatenated to the resulting vector of the convolutions consisting of 2,048 values. The result is passed to a fully connected portion with only two fully connected layers, one hidden and one output layer.
The hidden layer has 1,024 neurons and uses leaky ReLU activation with $\alpha = 0.1$. Since there are 14 body parts, the output layer has 14 neurons with linear activation. Since body part volumes are strongly influenced by body height and are sensitive to small changes, batch normalization was not used to preserve exact body height.\par
We calculated the \ac{MSE} between the ground truth and predicted body part volumes as loss since it automatically gives greater weight to large and influential errors. Such errors usually occur with greater frequency when estimating the volume of large body parts like the torso.
Also in this case we use Adam with default parameters and a learning rate of $\alpha = 2.5*10^{-4}$.

\section{Evaluation}
\label{ssec:Evaluation}
\noindent
\ac{MAPE} is used for evaluating the performance of the model. Compared to the \ac{MAE}, \ac{MAPE} has the advantage that small errors are weighted heavier in low volume models than high volume models. In other words, a total error of $5\mskip3mu\text{dm}^3$ is too much for a $40\mskip3mu\text{dm}^3$ person, while it is a good result for a $130\mskip3mu\text{dm}^3$ person.

\subsection{Basic Algorithm}
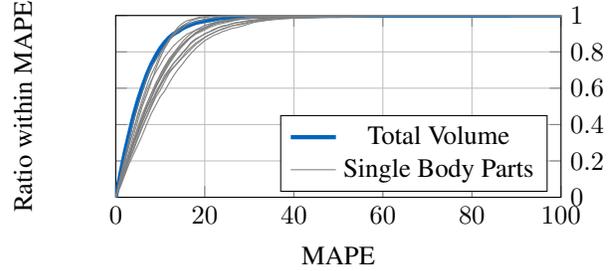
\begin{figure}
	\centering
	\begin{tikzpicture}
		\begin{axis}[
			legend pos=south east,
			grid=major,
			xlabel={\ac{MAPE}},
			ylabel={Ratio within \ac{MAPE}},
			yticklabel pos=right,
			width=0.9\linewidth,
			height=4cm,
			xmin = 0,
			xmax = 100,
			ymin = 0,
			ymax = 1,
			]
			\addplot[ultra thick, mark=none, color=Pantone300C] table [x=error, y=total_vol, col sep=comma]{data/results_best/cum_error_test.csv};
			\addlegendentry{Total Volume}
			
			\addplot[color=MediumGray, mark=none] table [x=error, y=head, col sep=comma]{data/results_best/cum_error_test.csv};
			\addplot[color=MediumGray, mark=none, forget plot] table [x=error, y=torso, col sep=comma]{data/results_best/cum_error_test.csv};
			\addplot[color=MediumGray, mark=none, forget plot] table [x=error, y=left_upper_arm, col sep=comma]{data/results_best/cum_error_test.csv};
			\addplot[color=MediumGray, mark=none, forget plot] table [x=error, y=left_fore_arm, col sep=comma]{data/results_best/cum_error_test.csv};
			\addplot[color=MediumGray, mark=none, forget plot] table [x=error, y=left_hand, col sep=comma]{data/results_best/cum_error_test.csv};
			\addplot[color=MediumGray, mark=none, forget plot] table [x=error, y=right_upper_arm, col sep=comma]{data/results_best/cum_error_test.csv};
			\addplot[color=MediumGray, mark=none, forget plot] table [x=error, y=right_fore_arm, col sep=comma]{data/results_best/cum_error_test.csv};
			\addplot[color=MediumGray, mark=none, forget plot] table [x=error, y=right_hand, col sep=comma]{data/results_best/cum_error_test.csv};
			\addplot[color=MediumGray, mark=none, forget plot] table [x=error, y=left_up_leg, col sep=comma]{data/results_best/cum_error_test.csv};
			\addplot[color=MediumGray, mark=none, forget plot] table [x=error, y=left_lower_leg, col sep=comma]{data/results_best/cum_error_test.csv};
			\addplot[color=MediumGray, mark=none, forget plot] table [x=error, y=left_foot, col sep=comma]{data/results_best/cum_error_test.csv};
			\addplot[color=MediumGray, mark=none, forget plot] table [x=error, y=right_up_leg, col sep=comma]{data/results_best/cum_error_test.csv};
			\addplot[color=MediumGray, mark=none, forget plot] table [x=error, y=right_lower_leg, col sep=comma]{data/results_best/cum_error_test.csv};
			\addplot[color=MediumGray, mark=none, forget plot] table [x=error, y=right_foot, col sep=comma]{data/results_best/cum_error_test.csv};
			\addlegendentry{Single Body Parts}
			
		\end{axis}
	\end{tikzpicture}
	\caption{Cumulative error of VolNet with the volume regressor on the test set.}
	\label{fig:VolNetCumError}
\end{figure}

\noindent
Figure \ref{fig:VolNetCumError} shows the cumulative error, meaning the ratio of predictions on the test set within a certain \ac{MAPE}. In Table \ref{tab:VolNetResultsValSet} the \ac{AE} and \ac{APE} of the model are given with their mean and standard deviation for the validation set. Using the validation set and measured in \ac{MAPE}, the model performed best on the total volume. Possibly because estimation errors of different body parts tend to cancel each other out.\par

\begin{table}[]
	\centering
	\begin{tabular}{@{}l|rr|rr@{}}
		\toprule
		& \multicolumn{2}{c|}{AE [dm\textsuperscript{3}]} & \multicolumn{2}{c}{APE [\%]} \\
		& $\mu$                     & $\sigma$                  & $\mu$    		& $\sigma$ \\
		\midrule
		head                  & 0.4           & 0.4           & 6.2                      & 5.2          \\
		torso                 & 3.0           & 3.2           & 7.8                      & 8.1          \\
		left upper arm        & 0.2           & 0.2           & 11.3                     & 10.3         \\
		left fore arm         & 0.1           & 0.1           & 8.6                      & 7.1          \\
		left hand             & 0.1           & 0.0           & 10.0                     & 8.4          \\
		right upper arm       & 0.2           & 0.2           & 10.4                     & 10.2         \\
		right fore arm        & 0.1           & 0.1           & 8.6                      & 7.5          \\
		right hand            & 0.1           & 0.0           & 10.0                     & 6.8          \\
		left up leg           & 0.5           & 0.5           & 9.8                      & 9.3          \\
		left lower leg        & 0.2           & 0.2           & 8.7                      & 7.1          \\
		left foot             & 0.1           & 0.1           & 6.8                      & 4.7          \\
		right up leg          & 0.5           & 0.5           & 9.7                      & 9.3          \\
		right lower leg       & 0.2           & 0.2           & 8.3                      & 7.0          \\
		right foot            & 0.1           & 0.1           & 6.4                      & 4.6          \\
		\textbf{total volume} & \textbf{4.7}  & \textbf{4.9}  & \textbf{6.2}             & \textbf{6.2} \\ \bottomrule
	\end{tabular}
	\caption{Performance of VolNet on the test set clearly showing that our architecture is able to predict the subjects body part volumes accurately.}
	\label{tab:VolNetResultsValSet}
\end{table}

On the test set, we obtained a 6.17\% \ac{MAPE} for the total volume, which is just slightly worse than the 5.3\% achieved on the validation set. Again, the \ac{MAPE} was the highest for the two arms (11.3\% and 10.4\%) and the lowest for the head (6.2\%). The results for all body parts are shown in Table \ref{tab:VolNetResultsValSet}.
VolNet correctly predicted $\sim$82\% of cases using a tolerance threshold equivalent to the 10\% of the body volume. Comparatively, BodyNet only correctly predicted $\sim$38\% of cases. Two examples where the model performed well are shown in the first and second column of Figure \ref{fig:results_examples_case}. For body part volumes both the prediction and the ground truth are given. The predictions for each single body part were accurate and resulted in good predictions for the total volume.
VolNet performed well on poses where the person is standing, walking or jumping, and with different arm positions. These are typical sport poses which are well represented in the training set. However, the network did not perform well with some challenging poses, for example, images where the person is shown from above or below. Intermediate tasks, however, provided better results but were still unable to accurately predict volumes. Therefore, some body poses have less information about the person's shape, and thus the volume, to allow the network to perform well. \par
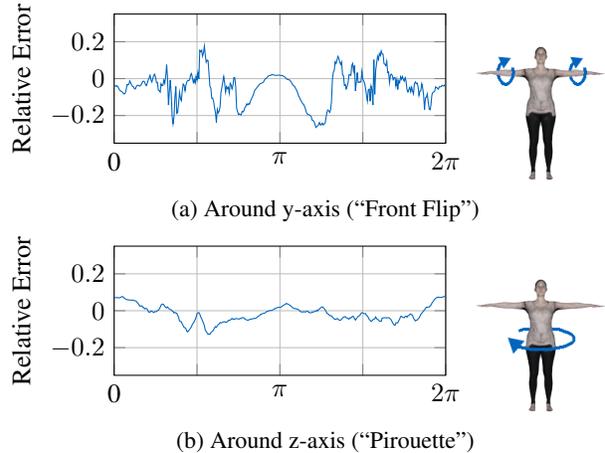
\begin{figure}[tb]
\centering
\begin{subfigure}{.49\textwidth}

  \begin{tikzpicture}
    \begin{axis}[
        ylabel={Relative Error},
		grid=major,
        scaled y ticks = false,
        width=0.7\linewidth,
        height=3.3cm,
        xmin = 0,
        xmax = 360,
        ymin = -0.35,
        ymax = 0.35,
        xtick={0,90,180,270,360},
        xticklabels={0,,$\pi$,,$2 \pi$},
        scaled x ticks = false,
    ]
    \addplot[solid, mark=none, color=Pantone300C] table [x=deg, y=PE, col sep=comma]{data/rotation_test/rotY.csv};
    \end{axis}
    \end{tikzpicture}
    \begin{tikzpicture}

    {\includegraphics[width=.2\textwidth]{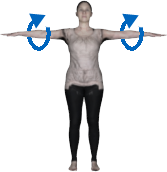}};
    
    \end{tikzpicture}
  \caption{Around y-axis (``Front Flip'')}
  \label{fig:results_rotation_test_Y}
\end{subfigure}
\begin{subfigure}{.49\textwidth}

  \begin{tikzpicture}
    \begin{axis}[
        ylabel={Relative Error},
		grid=major,
        scaled y ticks = false,
        width=0.7\linewidth,
        height=3.3cm,
        xmin = 0,
        xmax = 360,
        ymin = -0.35,
        ymax = 0.35,
        xtick={0,90,180,270,360},
        xticklabels={0,,$\pi$,,$2 \pi$},
        scaled x ticks = false,
    ]
	
    \addplot[solid, mark=none, color=Pantone300C] table [x=deg, y=PE, col sep=comma]{data/rotation_test/rotZ.csv};
    \end{axis}
    \end{tikzpicture}
    \begin{tikzpicture}
    {\includegraphics[width=.20\textwidth]{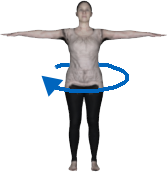}};
   
    \end{tikzpicture}
  \caption{Around z-axis (``Pirouette'')}
  \label{fig:results_rotation_test_Z}
\end{subfigure}%
\caption{Volume error in percent for rotating the person around its own axes. $\pi$ is always the neutral position of a person standing straight facing the camera.}
\label{fig:results_rotation_test}
\end{figure}

\par
To get a better understanding of how the pose influences the output of the model, different rotations are tested where a person in a neutral pose is rotated. We defined the neutral pose as the pose where a person is standing straight with arms outstretched facing the camera. First, the person is turned around the \textit{y-axis} producing a sort of front flip. Next, the person is turned around the \textit{z-axis} like for a pirouette. For each axis, 360$^{\circ}$ images with a constant interval of $1^{\circ}$ are generated and passed through the network.
The percentage error was calculated and the results are shown in Figure \ref{fig:results_rotation_test}. For rotation around the \textit{y-axis}, we noticed that the prediction for a person standing straight at $\pi$ works with very little error. Then, a small rotation in both direction leads to an error of up to -26\% which means that the model overestimates the person's volume. A small change in perspective up or down makes the person measured in pixels smaller within the photo. Since the real body height of the person is known and therefore constant, the person becomes wider from the perspective of the model and an overestimation of the volume. At some point around $\pi \pm \frac{\pi}{4}$ the model captures the change in perspective that translate into better accuracy.
At $\frac{\pi}{2}$ and $3\frac{\pi}{2}$ the person is only visible from below and above. In these positions, the estimation is inaccurate due to insufficient information. Predictions for the rotation around the \textit{z-axis }are more stable. Again, at $\pi$ the person looks straight into the camera. Even the predictions for the side views at $\frac{\pi}{2}$ and $3\frac{\pi}{2}$ are good. When the person is only visible from the back at around $0$ and $2\pi$, predictions are less stable. From this perspective, the body parts with high variance, such as the chest and belly, are not visible. Consequently, the model has too little information to correctly estimate the volume.

\subsection{Ablation Studies}
We studied the influence of each subtask on the final result by
training the last subnetwork responsible for regressing the volume with only a subset of inputs. For every combination of inputs, the last subnetwork was trained for five epochs and the best weights were used for testing. Results are reported in Table \ref{tab:ablation_study}. Increasing the number of inputs helped the network capture more details about the images. In fact, as seen in the last row, body height is crucial for network's performance.
In practice, body height could be obtained by using well-known keypoints from the surroundings. Furthermore, every subtask improved the overall performance. As seen in the second row of Table \ref{tab:ablation_study}, the 2D pose is essential for the network to capture something meaningful. The absolute improvement by adding the segmentation and 3D pose was marginal. However, the relative improvement was around 15\% for each additional input. Since the network was not trained end-to-end, and the loss over the volumes did not affect the first subnetworks, the improved performance is probably not achieved by a higher network capacity.

\begin{table}[]
\begin{tabular}{lr}
\textbf{Inputs to last subnetwork}                        &\textbf{ \ac{MAPE}}  \\
RGB, Body Height                                 & 26.25 \\ 
RGB, 2D Pose, Body Height                        & 8.45  \\ 
RGB, 2D Pose, Segm, Body Height                  & 7.25  \\ 
RGB, 2D Pose, Segm, 3D Pose, Body Height         & 6.17  \\ 
RGB, 2D Pose, Segm, 3D Pose                      &  13.37  
\end{tabular}
\caption{Ablation studies to measure the performance of different combinations of intermediate tasks.}
\label{tab:ablation_study}
\end{table}

\subsection{Single Body Parts}
We studied the volume estimation of a single body part. So far, the network is trained to predict all volumes at the same time. We investigated if performance could be improved by training the network to only learn the volume of a single body part.
Therefore, the best performing weights of all subnetworks were used and we fine-tuned the subnetwork responsible for regressing the volumes. For the evaluation, we used the left upper arm, because VolNet performed worst on this body part, and the head, because this might be of particular interest in many use-cases. \ac{MSE} was used as the loss function, but in this case it is only calculated for the single body part under consideration. The same parameters for the optimizer were used except that the learning rate was lowered to $1 \cdot 10^{-4}$.\par
The \ac{MAPE} for the head was improved from 6.3\% to 4.0\% and the left upper arm improved two percentage points to 8.1\%. When training on all body parts jointly, the network focuses more on the body parts that contribute the most to the total volume. Since \ac{MSE} is used as a loss function, the impact of larger body parts increases exponentially. Additionally, it might also improve as more capacity is available for only a single body part.

\section{Conclusion}
\label{ssec:conclusion}
\noindent
In this paper, we introduced VolNet: a deep neural network architecture able to predict human body volumes from a single 2D RGB image. To the best of our knowledge, this is the the first time such an approach has been used to accurately estimate body volumes from single RGB images and body height only.
We extended the SURREAL dataset with human body volumes leading to a novel dataset called SURREALvols. Using this dataset, we were able to estimate the total volume of a person, as well as the volumes of 14 single body parts with \textbf{$\sim$97\%} accuracy.
Our approach outperforms volume regression based on BodyNet by more than \textbf{$\sim$27\%} and sets a new baseline for body volume estimation. In addition, volume regression for single body parts can be further improved by up to \textbf{$\sim$2\%} when fine-tuning the network to specific body parts.

{\small
\bibliographystyle{ieee_fullname}
\bibliography{egpaper_final}

\begin{thebibliography}{10}\itemsep=-1pt

\bibitem{mocap}
Carnegie-mellon mocap database.
\newblock \url{http://mocap.cs.cmu.edu/}.
\newblock Accessed: 2020-09-30.

\bibitem{Affuso2018}
Olivia Affuso, Ligaj Pradhan, Chengcui Zhang, Song Gao, Howard~W. Wiener,
  Barbara Gower, Steven~B. Heymsfield, and David~B. Allison.
\newblock A method for measuring human body composition using digital images.
\newblock {\em PLOS ONE}, 13(11):1--13, 11 2018.

\bibitem{Hippocrate2016}
Elder~Akpa Akpro~Hippocrate, Hirohiko Suwa, Yutaka Arakawa, and Keiichi
  Yasumoto.
\newblock Food weight estimation using smartphone and cutlery.
\newblock In {\em Proceedings of the First Workshop on IoT-Enabled Healthcare
  and Wellness Technologies and Systems}, IoT of Health '16, page 9–14, New
  York, NY, USA, 2016. Association for Computing Machinery.

\bibitem{alldieck2018detailed}
Thiemo Alldieck, Marcus Magnor, Weipeng Xu, Christian Theobalt, and Gerard
  Pons-Moll.
\newblock Detailed human avatars from monocular video.
\newblock In {\em International Conference on 3D Vision}, pages 98--109, Sep
  2018.

\bibitem{alldieck2019tex2shape}
Thiemo Alldieck, Gerard Pons-Moll, Christian Theobalt, and Marcus Magnor.
\newblock Tex2shape: Detailed full human body geometry from a single image.
\newblock In {\em {IEEE} International Conference on Computer Vision ({ICCV})},
  2019.

\bibitem{Andriluka_2014_CVPR}
Mykhaylo Andriluka, Leonid Pishchulin, Peter Gehler, and Bernt Schiele.
\newblock 2d human pose estimation: New benchmark and state of the art
  analysis.
\newblock In {\em Proceedings of the IEEE Conference on Computer Vision and
  Pattern Recognition (CVPR)}, June 2014.

\bibitem{Badler1997}
N. {Badler}.
\newblock Virtual humans for animation, ergonomics, and simulation.
\newblock In {\em Proceedings IEEE Nonrigid and Articulated Motion Workshop},
  pages 28--36, 1997.

\bibitem{BenAbdelkaderStatBodyHeight}
C. {BenAbdelkader} and Y. {Yacoob}.
\newblock Statistical body height estimation from a single image.
\newblock In {\em 2008 8th IEEE International Conference on Automatic Face
  Gesture Recognition}, pages 1--7, 2008.

\bibitem{Bogo_ECCV_2016}
Federica Bogo, Angjoo Kanazawa, Christoph Lassner, Peter Gehler, Javier Romero,
  and Michael~J. Black.
\newblock Keep it {SMPL}: Automatic estimation of {3D} human pose and shape
  from a single image.
\newblock In {\em Computer Vision -- ECCV 2016}, Lecture Notes in Computer
  Science, pages 561--578. Springer International Publishing, Oct. 2016.

\bibitem{BogoSMPLify2016}
Federica Bogo, Angjoo Kanazawa, Christoph Lassner, Peter Gehler, Javier Romero,
  and Michael~J. Black.
\newblock Keep it {SMPL}: Automatic estimation of {3D} human pose and shape
  from a single image.
\newblock In {\em Computer Vision -- ECCV 2016}, Lecture Notes in Computer
  Science. Springer International Publishing, Oct. 2016.

\bibitem{Chen_2020_CVPR}
Zhiqin Chen, Andrea Tagliasacchi, and Hao Zhang.
\newblock {BSP-Net: Generating Compact Meshes via Binary Space Partitioning}.
\newblock In {\em IEEE/CVF Conference on Computer Vision and Pattern
  Recognition (CVPR)}, pages 42--51, 2020.

\bibitem{dantcheva2018}
Antitza Dantcheva, Fran{\c c}ois Bremond, and Piotr Bilinski.
\newblock {Show me your face and I will tell you your height, weight and body
  mass index}.
\newblock In {\em {International Coference on Pattern Recognition (ICPR)}},
  Beijing, China, Aug. 2018.

\bibitem{nhanes}
Center for Disease~Control and Prevention.
\newblock National health and nutrition examination survey.
\newblock \url{https://www.cdc.gov/nchs/nhanes/about_nhanes.htm}, 1999-2005.
\newblock Accessed: 2019-11-04.

\bibitem{Gabeur_2019_ICCV}
Valentin Gabeur, Jean-Sebastien Franco, Xavier Martin, Cordelia Schmid, and
  Gregory Rogez.
\newblock Moulding humans: Non-parametric 3d human shape estimation from single
  images.
\newblock In {\em Proceedings of the IEEE/CVF International Conference on
  Computer Vision (ICCV)}, October 2019.

\bibitem{guler2018densepose}
R{\i}za~Alp G{\"u}ler, Natalia Neverova, and Iasonas Kokkinos.
\newblock Densepose: Dense human pose estimation in the wild.
\newblock In {\em Proceedings of the IEEE Conference on Computer Vision and
  Pattern Recognition}, pages 7297--7306, 2018.

\bibitem{Gueler_2018_CVPR}
Rıza~Alp Güler, Natalia Neverova, and Iasonas Kokkinos.
\newblock Densepose: Dense human pose estimation in the wild.
\newblock In {\em Proceedings of the IEEE Conference on Computer Vision and
  Pattern Recognition (CVPR)}, June 2018.

\bibitem{Habermann_2020_CVPR}
Marc Habermann, Weipeng Xu, Michael Zollhofer, Gerard Pons-Moll, and Christian
  Theobalt.
\newblock {DeepCap: Monocular Human Performance Capture Using Weak
  Supervision}.
\newblock In {\em IEEE/CVF Conference on Computer Vision and Pattern
  Recognition (CVPR)}, pages 5051--5062, 2020.

\bibitem{ioffe2015batch}
Sergey Ioffe and Christian Szegedy.
\newblock Batch normalization: Accelerating deep network training by reducing
  internal covariate shift.
\newblock In {\em Proceedings of the 32nd International Conference on
  International Conference on Machine Learning - Volume 37}, ICML'15, page
  448–456. JMLR.org, 2015.

\bibitem{Jia2012}
W. {Jia}, Y. {Yue}, J.~D. {Fernstrom}, Z. {Zhang}, Y. {Yang}, and M. {Sun}.
\newblock 3d localization of circular feature in 2d image and application to
  food volume estimation.
\newblock In {\em 2012 Annual International Conference of the IEEE Engineering
  in Medicine and Biology Society}, pages 4545--4548, 2012.

\bibitem{Kingma_2014_adam}
Diederik Kingma and Jimmy Ba.
\newblock Adam: A method for stochastic optimization.
\newblock {\em International Conference on Learning Representations}, 12 2014.

\bibitem{Kolotouros_2019_ICCV}
Nikos Kolotouros, Georgios Pavlakos, Michael~J. Black, and Kostas Daniilidis.
\newblock Learning to reconstruct 3d human pose and shape via model-fitting in
  the loop.
\newblock In {\em Proceedings of the IEEE/CVF International Conference on
  Computer Vision (ICCV)}, October 2019.

\bibitem{LassnerUP2017}
Christoph Lassner, Javier Romero, Martin Kiefel, Federica Bogo, Michael~J.
  Black, and Peter~V. Gehler.
\newblock Unite the people: Closing the loop between 3d and 2d human
  representations.
\newblock In {\em IEEE Conf. on Computer Vision and Pattern Recognition
  (CVPR)}, July 2017.

\bibitem{Augmented_Reality2010}
Huei-Yung Lin and Ting-Wen Chen.
\newblock Augmented reality with human body interaction based on monocular 3d
  pose estimation.
\newblock In Jacques Blanc-Talon, Don Bone, Wilfried Philips, Dan Popescu, and
  Paul Scheunders, editors, {\em Advanced Concepts for Intelligent Vision
  Systems}, pages 321--331, Berlin, Heidelberg, 2010. Springer Berlin
  Heidelberg.

\bibitem{SMPL2015}
Matthew Loper, Naureen Mahmood, Javier Romero, Gerard Pons-Moll, and Michael~J.
  Black.
\newblock {SMPL}: A skinned multi-person linear model.
\newblock {\em ACM Trans. Graphics (Proc. SIGGRAPH Asia)}, 34(6):248:1--248:16,
  Oct. 2015.

\bibitem{Luo_2020_CVPR}
Keyang Luo, Tao Guan, Lili Ju, Yuesong Wang, Zhuo Chen, and Yawei Luo.
\newblock {Attention-Aware Multi-View Stereo}.
\newblock In {\em Proceedings of the IEEE/CVF Conference on Computer Vision and
  Pattern Recognition (CVPR)}, 2020.

\bibitem{moreno2018noiseadjustment}
F.~J. {Moreno-Barea}, F. {Strazzera}, J.~M. {Jerez}, D. {Urda}, and L.
  {Franco}.
\newblock Forward noise adjustment scheme for data augmentation.
\newblock In {\em 2018 IEEE Symposium Series on Computational Intelligence
  (SSCI)}, pages 728--734, 2018.

\bibitem{Mori2006}
G. {Mori} and J. {Malik}.
\newblock Recovering 3d human body configurations using shape contexts.
\newblock {\em IEEE Transactions on Pattern Analysis and Machine Intelligence},
  28(7):1052--1062, 2006.

\bibitem{pmlr-v28-muandet13}
Krikamol Muandet, David Balduzzi, and Bernhard Schölkopf.
\newblock Domain generalization via invariant feature representation.
\newblock {\em ICML}, 01 2013.

\bibitem{navaneet2019differ}
K~L Navaneet, Priyanka Mandikal, Varun Jampani, and R~Venkatesh Babu.
\newblock {DIFFER}: Moving beyond 3d reconstruction with differentiable feature
  rendering.
\newblock In {\em CVPR Workshops}, 2019.

\bibitem{newell2016stacked}
Alejandro Newell, Kaiyu Yang, and Jia Deng.
\newblock Stacked hourglass networks for human pose estimation.
\newblock In {\em ECCV}, 2016.

\bibitem{Pavlakos_2018_CVPR}
Georgios Pavlakos, Luyang Zhu, Xiaowei Zhou, and Kostas Daniilidis.
\newblock Learning to estimate 3d human pose and shape from a single color
  image.
\newblock In {\em Proceedings of the IEEE Conference on Computer Vision and
  Pattern Recognition (CVPR)}, June 2018.

\bibitem{Robinette_caesar_2002}
Kathleen Robinette, Sherri Blackwell, Hein Daanen, Mark Boehmer, and Scott
  Fleming.
\newblock Civilian american and european surface anthropometry resource
  (caesar), final report. volume 1. summary.
\newblock page~74, 06 2002.

\bibitem{Shorten2019ASO}
Connor Shorten and T. Khoshgoftaar.
\newblock A survey on image data augmentation for deep learning.
\newblock {\em Journal of Big Data}, 6:1--48, 2019.

\bibitem{Thalmann1996}
D. {Thalmann}, {Jianhua Shen}, and E. {Chauvineau}.
\newblock Fast realistic human body deformations for animation and vr
  applications.
\newblock In {\em Proceedings of CG International '96}, pages 166--174, 1996.

\bibitem{Tong2012}
J. {Tong}, J. {Zhou}, L. {Liu}, Z. {Pan}, and H. {Yan}.
\newblock Scanning 3d full human bodies using kinects.
\newblock {\em IEEE Transactions on Visualization and Computer Graphics},
  18(4):643--650, 2012.

\bibitem{Toshev_2014_CVPR}
Alexander Toshev and Christian Szegedy.
\newblock Deeppose: Human pose estimation via deep neural networks.
\newblock In {\em Proceedings of the IEEE Conference on Computer Vision and
  Pattern Recognition (CVPR)}, June 2014.

\bibitem{varol18_bodynet}
G{\"u}l Varol, Duygu Ceylan, Bryan Russell, Jimei Yang, Ersin Yumer, Ivan
  Laptev, and Cordelia Schmid.
\newblock {BodyNet}: Volumetric inference of {3D} human body shapes.
\newblock In {\em ECCV}, 2018.

\bibitem{varol17_surreal}
G{\"u}l Varol, Javier Romero, Xavier Martin, Naureen Mahmood, Michael~J. Black,
  Ivan Laptev, and Cordelia Schmid.
\newblock Learning from synthetic humans.
\newblock In {\em Proceedings IEEE Conference on Computer Vision and Pattern
  Recognition (CVPR) 2017}, Piscataway, NJ, USA, July 2017. IEEE.

\bibitem{velardo2010weight}
Carmelo Velardo and Jean-Luc Dugelay.
\newblock Weight estimation from visual body appearance.
\newblock In {\em 2010 Fourth IEEE International Conference on Biometrics:
  Theory, Applications and Systems (BTAS)}, pages 1--6. IEEE, 2010.

\bibitem{Wu_2020_CVPR}
Shangzhe Wu, Christian Rupprecht, and Andrea Vedaldi.
\newblock {Unsupervised Learning of Probably Symmetric Deformable 3D Objects
  From Images in the Wild}.
\newblock In {\em Proceedings of the IEEE/CVF Conference on Computer Vision and
  Pattern Recognition (CVPR)}, pages 1--10, 2020.

\bibitem{xie2016aggregated}
S. {Xie}, R. {Girshick}, P. {Dollár}, Z. {Tu}, and K. {He}.
\newblock Aggregated residual transformations for deep neural networks.
\newblock In {\em 2017 IEEE Conference on Computer Vision and Pattern
  Recognition (CVPR)}, pages 5987--5995, 2017.

\bibitem{Xu2013}
Chang Xu, Ye He, Nitin Khannan, Albert Parra, Carol Boushey, and Edward Delp.
\newblock Image-based food volume estimation.
\newblock In {\em Proceedings of the 5th International Workshop on Multimedia
  for Cooking \& Eating Activities}, CEA '13, page 75–80, New York, NY, USA,
  2013. Association for Computing Machinery.

\bibitem{Xu_2019_ICCV}
Yuanlu Xu, Song-Chun Zhu, and Tony Tung.
\newblock Denserac: Joint 3d pose and shape estimation by dense
  render-and-compare.
\newblock In {\em Proceedings of the IEEE/CVF International Conference on
  Computer Vision (ICCV)}, October 2019.

\bibitem{yang_2013_pck_metric}
Yi Yang and Deva Ramanan.
\newblock Articulated human detection with flexible mixtures of parts.
\newblock {\em IEEE transactions on pattern analysis and machine intelligence},
  35:2878--90, 12 2013.

\bibitem{Yu_LSUN_2015}
Fisher Yu, Yinda Zhang, Shuran Song, Ari Seff, and Jianxiong Xiao.
\newblock {LSUN:} construction of a large-scale image dataset using deep
  learning with humans in the loop.
\newblock {\em CoRR}, abs/1506.03365, 2015.

\bibitem{Zanfir_2018_CVPR}
Andrei Zanfir, Elisabeta Marinoiu, and Cristian Sminchisescu.
\newblock Monocular 3d pose and shape estimation of multiple people in natural
  scenes - the importance of multiple scene constraints.
\newblock In {\em Proceedings of the IEEE Conference on Computer Vision and
  Pattern Recognition (CVPR)}, June 2018.

\bibitem{Zhang_2017_CVPR}
Chao Zhang, Sergi Pujades, Michael~J. Black, and Gerard Pons-Moll.
\newblock Detailed, accurate, human shape estimation from clothed 3d scan
  sequences.
\newblock In {\em Proceedings of the IEEE Conference on Computer Vision and
  Pattern Recognition (CVPR)}, July 2017.

\end{thebibliography}
}

\end{document}